\documentclass[11pt,a4paper]{article}

\usepackage[utf8]{inputenc}
\usepackage[T1]{fontenc}
\usepackage{lmodern}
\usepackage[margin=2.5cm]{geometry}
\usepackage{amsmath,amssymb,amsfonts}
\usepackage{graphicx}
\usepackage{booktabs}
\usepackage[numbers,sort&compress]{natbib}
\usepackage{authblk}
\usepackage{xcolor}
\usepackage[colorlinks=true,allcolors=blue!60!black]{hyperref}
\usepackage{fancyvrb}
\usepackage{embedfile}

\newcommand{\fR}{f_{\mathrm{R}}}  

\embedfile[desc={BibTeX entry for this paper}]{gioia2025contactless.bib}

\title{Contactless estimation of respiratory frequency using 3D-CNN on thermal images}

\author[1]{Federica Gioia\thanks{Equal contribution.}}
\author[2]{Filippo Pura\protect\footnotemark[1]}
\author[1]{Alberto Greco}
\author[2]{Dario Piga}
\author[3,4]{Arcangelo Merla}
\author[2]{Marco Forgione}

\affil[1]{Department of Information Engineering, University of Pisa, Italy}
\affil[2]{SUPSI-DTI-IDSIA, Dalle Molle Institute for Artificial Intelligence, Lugano-Viganello, Switzerland}
\affil[3]{Department of Engineering and Geology, University ``G.~d'Annunzio'' Chieti-Pescara, Pescara, Italy}
\affil[4]{Next2U s.r.l., Pescara, Italy}

\date{}

\begin{document}

\maketitle

\begin{center}
\small\itshape
This is the accepted version of an article published in the IEEE Journal of Biomedical and Health Informatics, vol.~29, no.~10, pp.~7387--7395, 2025, DOI: \href{https://doi.org/10.1109/JBHI.2025.3567141}{10.1109/JBHI.2025.3567141}.\\
\textcopyright~2025 IEEE. Personal use of this material is permitted. Permission from IEEE must be obtained for all other uses, in any current or future media, including reprinting/republishing this material for advertising or promotional purposes, creating new collective works, for resale or redistribution to servers or lists, or reuse of any copyrighted component of this work in other works.
\end{center}

\noindent{\small\textbf{Please cite the published version:}}
\begin{Verbatim}
@article{gioia2025contactless,
  title   = {Contactless Estimation of Respiratory Frequency Using {3D-CNN} on Thermal Images},
  author  = {Gioia, Federica and Pura, Filippo and Greco, Alberto and
             Piga, Dario and Merla, Arcangelo and Forgione, Marco},
  journal = {IEEE Journal of Biomedical and Health Informatics},
  volume  = {29},
  number  = {10},
  pages   = {7387--7395},
  year    = {2025},
  doi     = {10.1109/JBHI.2025.3567141}
}
\end{Verbatim}

\begin{abstract}
Monitoring physiological parameters such as respiratory rate ($\fR$) is essential for diagnosing and managing various pathological conditions. Thermal imaging offers a promising contactless alternative to traditional methods, which often rely on partially invasive sensors or obtrusive wearable systems. However, existing approaches for $\fR$ estimation from thermal signals typically require extensive pre-processing and manual or semi-automatic region-of-interest (ROI) tracking, limiting their practical applicability.

This study proposes a deep learning-based method for estimating $\fR$ directly from thermal videos, eliminating the need for complex pre-processing and ROI tracking. A 3D Convolutional Neural Network (3D-CNN) is developed to operate on raw thermal video data. To address challenges related to small datasets, the model is trained using data augmentation and transfer learning from synthetic datasets.

Experimental results demonstrate that the proposed approach achieves a validation $R^2$ score of approximately 0.61 on both pre-processed and raw thermal videos. By simplifying the workflow, this method holds promise for enhancing the feasibility of thermal imaging in real-world applications, such as remote healthcare and driver monitoring in automotive applications.
\end{abstract}


\section{Introduction}
\label{sec:introduction}

Respiratory rate, also known as respiratory frequency ($\fR$), is a vital sign of significant importance sensitive to different pathological conditions and stressors.
Within the healthcare field, it is used for diagnosing, prognosing, and managing pneumonia \cite{goodman2019challenges}. Moreover, there is significant evidence indicating that $\fR$ serves as a crucial predictor for clinical deterioration and cardiac arrest \cite{massaroni2020contactless, nicolo2020importance}. In addition, the $\fR$ is an indicator in neonatal intensive care units \cite{maurya2023non, rolfe2019importance}. Indeed, respiratory diseases stand out as a primary cause of morbidity among neonates, with a higher prevalence among preterm infants \cite{pramanik2015neonatal}. Monitoring respiration, specifically $\fR$, aids in detecting sudden infant death syndrome, one of the major causes of death in infants \cite{kumar2020continuous, pereira2017estimation}.
In the field of psychophysiology, $\fR$ serves as a reliable indicator of mental stress and cognitive load. It proves valuable for assessing stress levels not only in workplace and computer-related tasks but also in the contexts of aviation and automotive to assess stress levels or drowsiness in pilots and drivers \cite{massaroni2020contactless, grassmann2016role, mateu2019non, massaroni2019non, siddiqui2021non}.
Finally, $\fR$ is a reliable indicator of physical effort and fatigue \cite{nicolo2020respiratory,nicolo2017respiratory}.

Nonetheless, $\fR$ remains one of the least documented and underestimated vital signs in clinical practice, mainly due to limitations in current clinical monitoring techniques \cite{elliott2016respiratory,nicolo2017respiratory,flenady2017accurate,loughlin2018respiratory, nicolo2020importance}. Indeed, manual counting of respiratory cycles is still performed in cardio-respiratory resuscitation, posing a challenging task even for trained professionals, and yielding subjective outcomes \cite{nicolo2020importance}. Other modalities rely on the attachment of sensors to the patient's body, which induces stress and discomfort. Indeed, sensor placement in psychophysiology has the limitation of potentially altering the natural psychophysiological behavior of the subject monitored. Furthermore, the removal of adhesive electrodes often leads to epidermal stripping and skin disruption in neonatal intensive care \cite{pereira2018noncontact}.
To overcome these limitations, contactless methods have been proposed to monitor respiration \cite{massaroni2020contactless}.
In particular, InfraRed Thermography (IRT) has been taking hold because of the great advantage of remotely monitoring the infrared radiation emitted by objects, enabling passive and non-intrusive monitoring of skin temperature.
Unlike visible recordings, IRT remains unaffected by lighting conditions and can function in the darkness, making it well-suited for monitoring drivers or conducting sleep studies. Notably, respiratory cycles are visible to the naked eye in IRT, as exhaled air, being warmer than inhaled air, generates observable variations in radiation intensity or temperature within the nasal/nostrils region of interest (ROI).

Therefore, most imaging methods for $\fR$ monitoring involve identifying and tracking specific features or anatomic landmarks to extract the respiration signal by averaging the temperature of the pixels within a chosen ROI, such as the nose or nostrils, over time. For instance,~\cite{fei2009thermistor} introduced a network of probabilistic trackers to segment the nostril region and exploited wavelet analysis to reconstruct the breathing information. A related approach was followed in Abbas et al.~\cite{abbas2011neonatal} for neonatal non-contact monitoring of the respiratory activity. The ROI was defined in the first frame and no tracking algorithm was used. Building on this, Pereira et al.~\cite{pereira2017estimation} manually selected the nose region in the first frame and applied a tracking algorithm to compensate for the motion in the following frames. In \cite{kiashari2020evaluation}, the authors proposed a method to automatically detect the respiration region using Canny edge detector.
The contribution outlined in \cite{jagadev2020non} consists of a real-time tracking of human nostrils using an ensemble of regression trees. The authors demonstrate its robustness in scenarios involving head motion and object occlusion. Other studies suggested to consider not only temperature modulation around the mouth and nostrils but also the body movements, detecting respiration by blending the temperature change of ROIs and breathing motion \cite{jakkaew2020non, barbosa2017estimation}. A multi-sensor approach, using both visible and infrared technologies, to increase the robustness and reliability of region detection and tracking has been used in \cite{barbosa2017estimation, hu2017synergetic}. Similarly, Scebba et al. fused the information obtained in the near-IR and far-IR spectra from the nostril and chest \cite{scebba2020multispectral}. They extracted multiple streams of respiratory information from independent video modalities and then applied two dedicated data fusion models to determine whether the subject was breathing, and at which frequency. Yet, a noteworthy limitation of these studies lies in the demand for sophisticated algorithms not only to detect and track specific regions of interest but also to accomplish frame registration across diverse modalities.
ROI detection to identify the nose in IR videos in order to extract breathing-relevant pixels can be particularly challenging if the videos have low contrast or suffer from suboptimal focusing. These conditions may compromise the overall robustness of the approach, making accurate ROI detection more challenging and potentially affecting the reliability of breathing signal extraction. To skip the region selection and tracking procedure, Scebba et al.~\cite{scebba2018ir} extracted respiration signals based on pixel time series, avoiding image segmentation. However, they cropped the frames around the nose, making the algorithm still region-based. A grid-based approach, which does not use any anatomical region selection, was proposed in \cite{pereira2018noncontact} to estimate $\fR$. A signal quality index was empirically developed to automatically distinguish ROI in the video containing the respiratory signal from those containing only noise. They used Bayesian fusion of the selected ROIs to get robust estimations for $\fR$. The major drawback of the proposed algorithm is the sensitivity to motion artifacts. Moreover, both ROI-based and grid-based approaches require adequate pre-processing of the recordings to ensure a reasonable signal to noise ratio of the extracted breathing signal.

In this study, a new approach to predict the $\fR$ from thermal videos using deep learning is proposed. Ultimately, our objective was to develop a model capable of operating directly on raw thermal videos, eliminating the need for additional pre-processing steps. In the literature, deep learning methods were used for face detection and face tracking, and to infer stress and drowsiness levels from thermal imaging of people's breathing patterns \cite{cho2017deepbreath, ebrahimian2022multi}.
However, to the best of the authors' knowledge, this is the first attempt to employ end-to-end deep learning for direct extraction of respiratory signals from thermal videos. With this work, the limitations of ROI-based $\fR$ extraction are addressed and the need for intense pre-processing of the thermal videos is removed.
Specifically, a 3D Convolutional Neural Network (3D-CNN) was employed to analyze the thermal video and predict the corresponding $\fR$. The model was trained and evaluated using $\fR$ data extracted through a standard protocol from the respiratory signal obtained via a chest belt, used as reference. In addition, an innovative and creative data augmentation procedure is introduced to address the constraints posed by small datasets, a common limitation of deep learning in healthcare applications. To validate the approach a Leave-Subject-Out strategy is adopted to avoid the bias due to the presence of the same subject in both training and test set. This approach enhanced the generalization capability of the model, allowing it to work effectively on previously unseen subjects. Finally, the investigation was extended to both processed and raw thermal videos. This allowed model performance comparison across multiple scenarios, ensuring a comprehensive assessment of its capabilities.

\section{Materials and Methods}

\subsection{Subjects Recruitment, Experimental Protocol and Acquisition Set-up}
Thirty healthy volunteers (age range of 26.6$\pm$3.6 years, 20 females), with no history of cardiovascular or mental disorders, participated in this study. The experiments were approved by the Bioethics Committee of the University of Pisa (no. 85 15/2019), and each volunteer signed the informed consent. The subjects were asked to avoid alcohol, caffeine, and physical exercise for 24 hours before the experimental session. They were also asked not to use make-up and/or moisturizers for the experiment and not to shave within 48 hours before the session.

During the experiment, standard procedures for IR imaging acquisition are followed \cite{Jones2012}. Participants were seated in a comfortable chair
in a controlled environment, far from ventilation and radiation sources. Temperature and humidity were kept in the range of approximately 24--26$^\circ$C and 40--60$\%$, respectively. Within the same experimental session, the temperature change never exceeded 1$^\circ$C and humidity variation was a maximum of 5$\%$. Before the start of the experiment, the subjects acclimatized for at least 10 minutes to the testing environment.

Facial thermograms were acquired using a FLIR T640 thermal camera with a 24.6 mm lens.
The camera features a Focal Plane Array (FPA) size of $640 \times 480$, corresponding to 307,200 uncooled microbolometer detectors, with no interpolation applied, Noise Equivalent Differential Temperature (NETD) $<$ 0.04 mK @ +30$^\circ$C and spectral range of 7.8--14 $\mu$m in the Long Wave InfraRed (LWIR). The sampling frequency was 5 Hz. The camera's emissivity was set to 0.98, the recommended value for a human body surface, and it was placed at a fixed distance of 1 meter from the face of the subject. Twenty minutes before the start of the recording, the camera was turned on to allow for sensor stabilization. Their respiratory signal was simultaneously acquired using a BIOPAC MP150 system, with a piezoelectric chest belt, at a sampling rate of 250 Hz.

\begin{figure}[t]
	\centering
	\includegraphics[width=\linewidth]{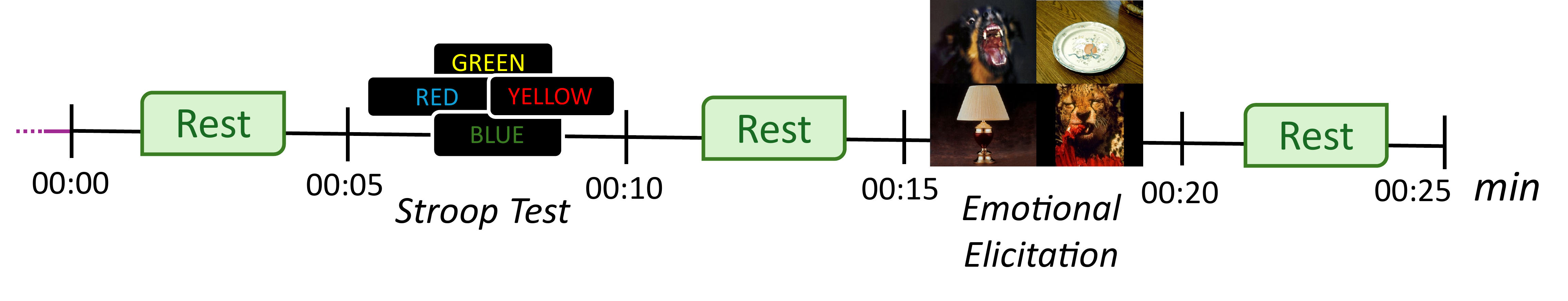}
	\caption{Experimental timeline showing the succession of 5-min tasks administered to each subject. After an initial acclimatization period (dashed line), participants underwent a baseline resting phase, followed by a Stroop test (cognitive stress induction), a second resting period, and an emotional elicitation task using affective images.}
	\label{Timeline}
\end{figure}

The experimental protocol, as illustrated in Fig.~\ref{Timeline}, was designed to alter the respiratory frequency of the subjects during the different 5-minute tasks.
Specifically, the protocol involved alternating between resting phases and mental and emotional stimulation. The first task involved a Stroop test \cite{stroop1935studies}, previously employed in stress classification based on respiratory frequency \cite{cho2017deepbreath}. This task required participants to rapidly solve incongruous stimuli, with a buzzer alerting them for each missed or incorrect answer, as detailed in \cite{gioia2023investigating}. The second task consisted of displaying a sequence of images, selected from the International Affective Picture System (IAPS) database, to the subjects. This public database is a consistent and effective resource for eliciting emotions \cite{lang2005international}. In particular, pictures with high arousal and negative valence were presented to the participant one after the other, for 5 seconds each.

Overall, the range of variability of respiratory frequency in the dataset is between 9 and 30 cycles per minute, see the evolution of $\fR$ over time for 5 test subjects in Fig.~\ref{PredTrue}.

\begin{figure}[t]
	\centering
	\includegraphics[width=0.6\linewidth]{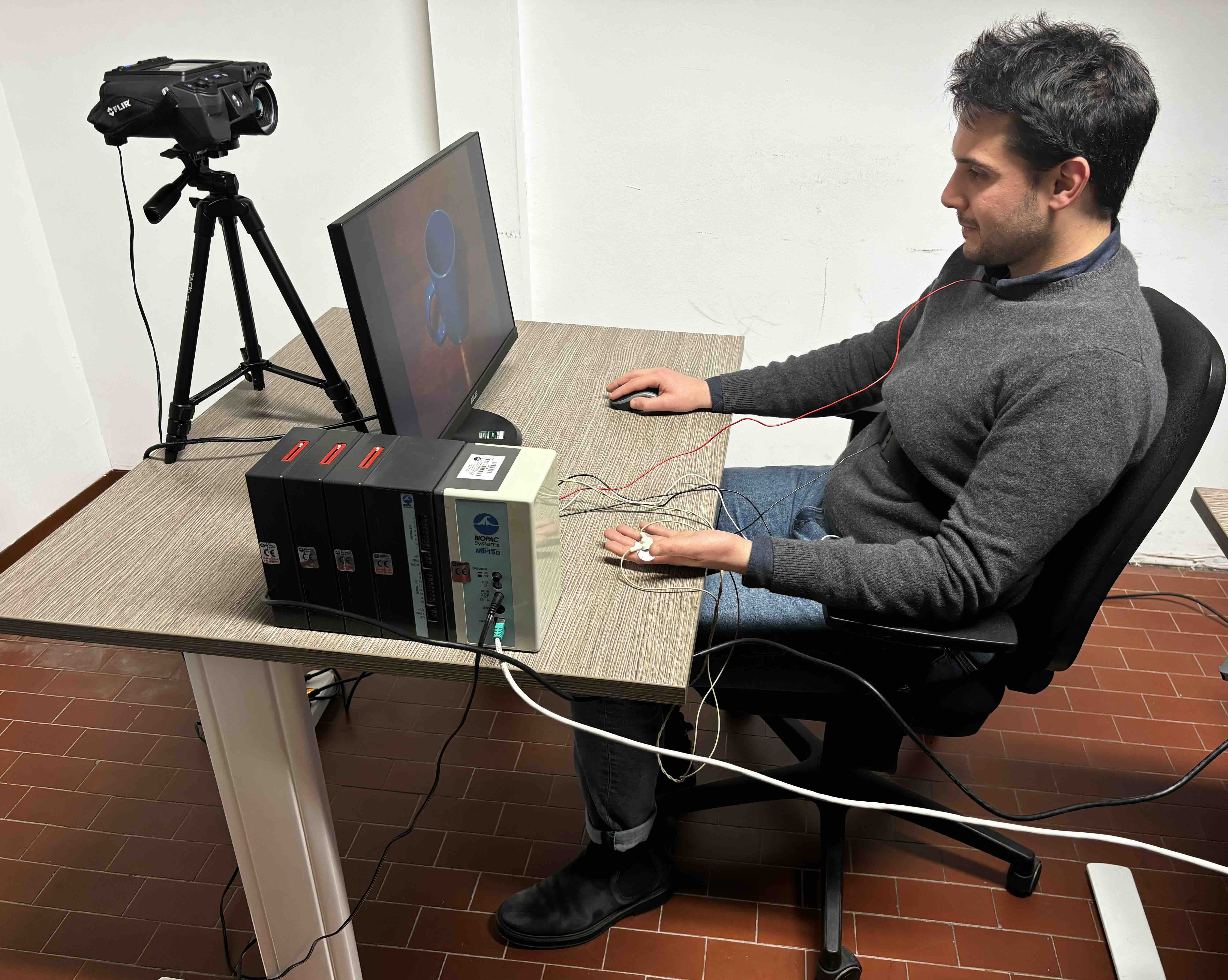}
	\caption{Overview of the experimental setup. The participant was seated in a thermally controlled room, facing a FLIR T640 thermal camera positioned at a 1-meter distance. The thermal camera acquired infrared facial images, while a chest belt connected to a BIOPAC MP150 system recorded the reference respiratory signal. The setup ensured synchronized acquisition of both contactless and ground-truth respiratory data.}
	\label{fig:setup}
\end{figure}

The experimental setup is visualized in Fig.~\ref{fig:setup}.
The interested reader is referred to~\cite{gioia2024thermica} for further details about the experimental setup, the protocol, and an exploratory analysis of the measured variables.

\subsection{Thermal Recording Preprocessing}
The thermal frames were coregistered to reduce motion using a dense optical flow algorithm as in \cite{sonkusare2021data}. All frames were spatially Gaussian smoothed to remove high-frequency noise. Moreover, the original IR frames were downsized to $60 \times 80$ pixels, effectively reducing their original size by a factor of 8, in order to lower the computational load.
\begin{figure}[t]
	\centering
	\includegraphics[width=0.75\linewidth]{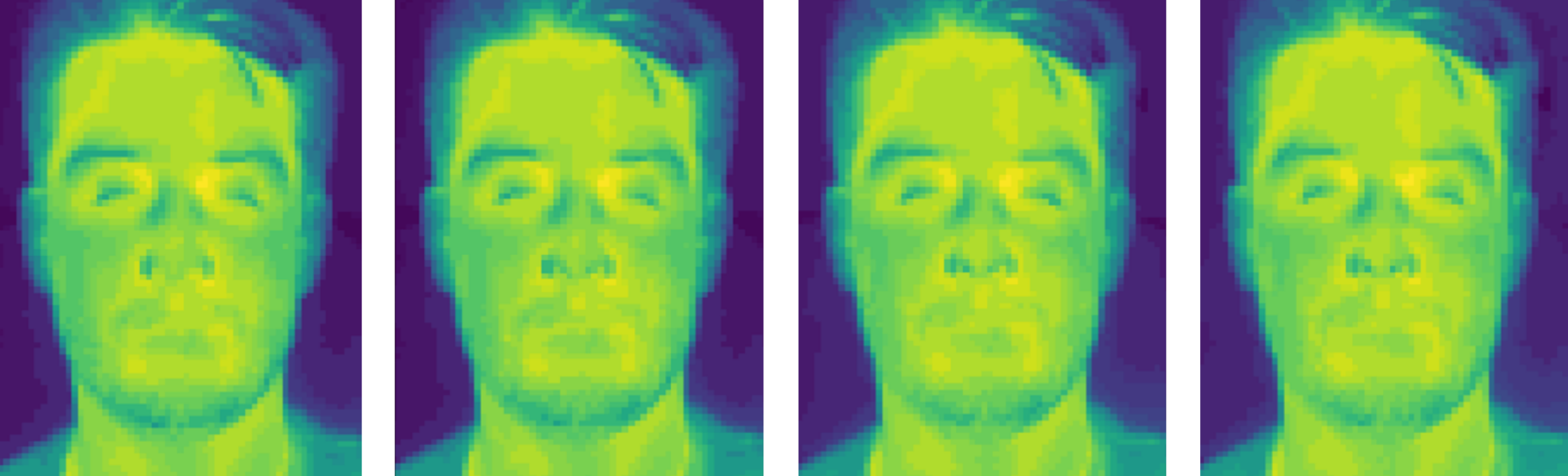}
	\caption{Sequence of four consecutive unprocessed IR frames downsampled to $60 \times 80$ pixels.}
	\label{IRframes}
\end{figure}
A mask to remove the background and the eyes was applied to all frames before the analysis. In particular, the eyes were excluded to remove blinking artifacts. The face segmentation mask was created by shaping an ellipse centered in the nose tip, with a minor axis length equal to the width of the face (being determined by the location of landmarks close to the ears) and a major axis length equal to twice the distance between the nose tip and the chin landmarks. Thermal and visible frames were registered using a Local Weighted Mean transformation, which outperforms other geometrical transformations \cite{cardone2021automated}. Facial landmarks were identified in the RGB frames using the Yuval Nirkin algorithm \cite{nirkin2018face}. Analogously, the ellipses to exclude the eyes were automatically determined based on the landmarks' location.
\begin{figure}[t]
	\centering
	\includegraphics[width=0.75\linewidth]{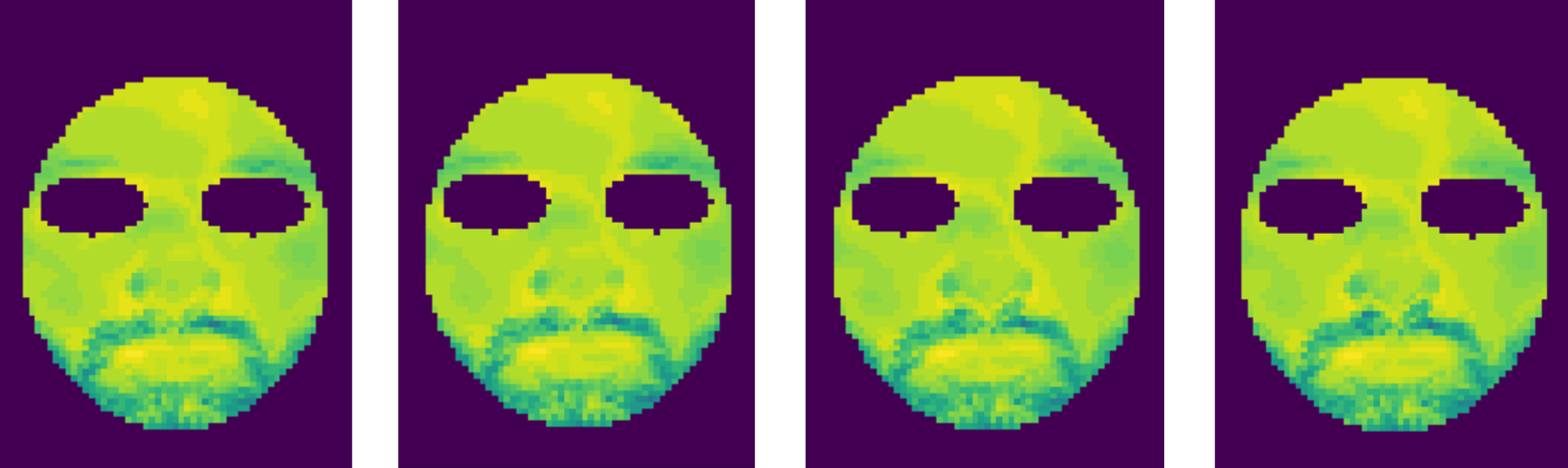}
	\caption{Sequence of four consecutive pre-processed IR frames. A mask is applied to suppress the background and the eye region.}
	\label{Preprocessed}
\end{figure}

In the paper, the respiratory frequency reconstruction algorithm will be assessed both on
the unprocessed IR videos (downsampled to $60 \times 80$ pixel to reduce the computational complexity) and on the IR videos pre-processed as described in the previous paragraphs, see Fig.~\ref{IRframes} and Fig.~\ref{Preprocessed}, respectively.

\subsection{Reference Respiratory Frequency}
\label{sec:resp_freq_extraction}
The respiratory signal was used to extract the respiratory frequency ($\fR$), which then served as the target output for training and evaluating the model. The identification of $\fR$ from the respiratory signal involved pinpointing the maximum peak within the
frequency spectrum, specifically within the range of 0.15 to 0.55 Hz.
The frequency spectrum was obtained using the Fast Fourier Transform, applied to a sliding window of 30-second long signals with 29 seconds of overlap. This approach aimed to capture the dynamic changes in $\fR$ over each 5-minute task period.

\subsection{Deep Learning - 3D-CNN}
\label{sec:3DCNN}
The 3D Convolutional Neural Network \cite{tran2015learning} is a deep-learning architecture designed to process three-dimensional data objects such as molecular structure (three spatial dimensions) and videos (one temporal and two spatial dimensions). Recently, this architecture has been successfully applied to predict the protein-ligand interaction \cite{jimenez2018k} and for human action recognition from videos \cite{3dcnnactionrecognition}.

\begin{figure}[t]
	\centering
	\includegraphics[width=\linewidth]{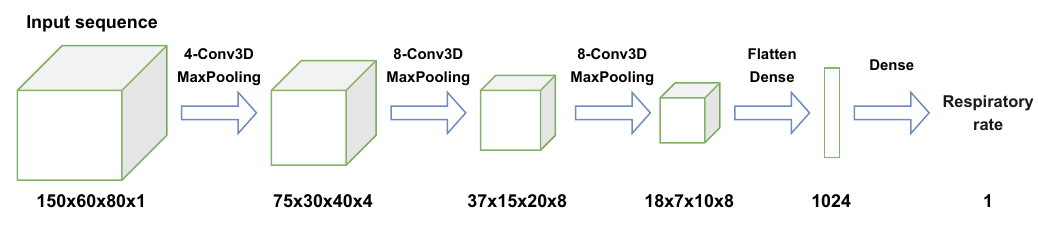}
	\caption{The 3D-CNN architecture used in this study. Each 3D convolutional layer employs a $3 \times 3 \times 3$ kernel with stride and padding set to $1$, while max pooling layers use a $2 \times 2 \times 2$ kernel. ReLU activation functions are applied after each convolutional and dense layer, except for the final dense layer with output dimension of $1$, which directly outputs the frequency prediction. The dimensions of the network's input, output, and intermediate activations are indicated below each corresponding block.}
	\label{Architecture}
\end{figure}

In this paper, a 3D-CNN is used to reconstruct the respiratory frequency $\fR$ from the measured IR video stream.
Overall, the network receives as input a \emph{tensor} of dimension $150 \times 60 \times 80$, corresponding to a 30-second video stream at sampling frequency of 5~Hz (thus, 150 frames) and resolution $60 \times 80$ pixels, and produces as output a single scalar value representing the mean respiratory frequency over the same time window.
The network architecture comprises three 3D convolutional layers interleaved with Max Pooling operations, followed by two dense layers, as illustrated in Fig.~\ref{Architecture}.

\begin{figure}[t]
	\centering
	\includegraphics[width=0.75\linewidth]{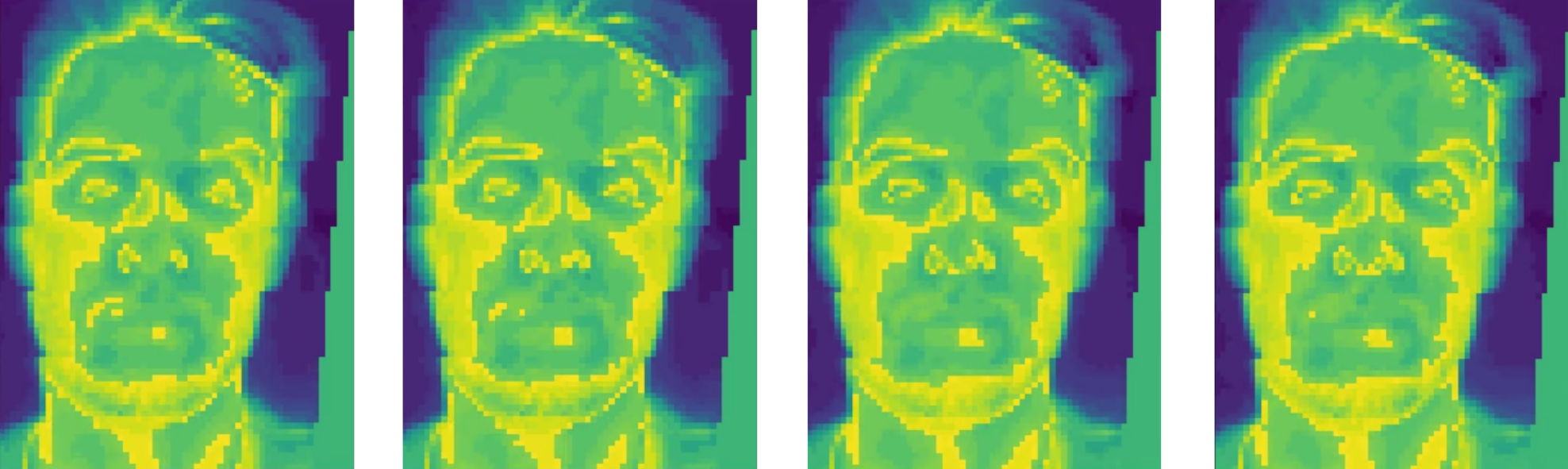}
	\caption{Augmented frames from the unprocessed IR dataset. The \texttt{RandAugment} algorithm is applied to the frames in Fig.~\ref{IRframes}.}
	\label{Augmented}
\end{figure}

Input and output instances for training are obtained by extracting sequences of 150 consecutive IR frames from the longer experiment videos, together with the mean respiratory frequencies in the corresponding 30-second periods.
A \emph{sliding window} of one single frame is used, i.e.\ all possible (partially overlapping) 150-length sequences are fed to the model during training.
Moreover, to cope with the relatively small size of the available dataset,
the extracted video segments are transformed by randomized \emph{data augmentation} operations that include noise corruption, blurring, distortions, and rotations, according to the \texttt{RandAugment} algorithm~\cite{randomaugment}, see Fig.~\ref{Augmented}.
These transformations are aimed at improving the generalization property of the trained model and tackle, at least in part, the
\emph{overfitting} issue commonly encountered when the training dataset lacks either quantity or diversity~\cite{xu2023comprehensive}.

\begin{figure}[t]
	\centering
	\includegraphics[width=0.75\linewidth]{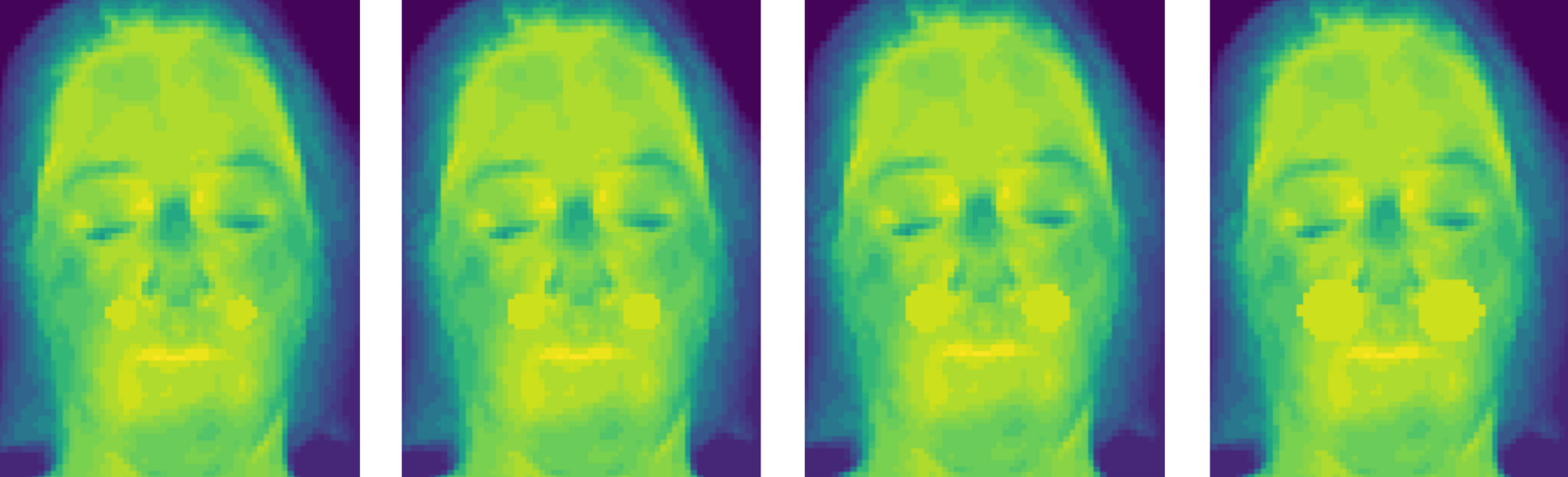}
	\caption{Frames extracted from the synthetic dataset. Pulsating circles around the nostril region simulate the respiratory activity.}
	\label{Synthetic}
\end{figure}

To further enhance generalization, a \emph{sim-to-real} transfer learning approach~\cite{kim2022transferable} from an initial model learned over a \emph{synthetic dataset} is followed. Specifically, synthetic video segments are generated by combining fixed facial backgrounds with a \emph{simulated} respiratory activity. The latter is implemented by superposing two pulsating circles with randomized position, size, and pulsation frequency to the fixed backgrounds, thereby mimicking the behavior of nostrils (see Fig.~\ref{Synthetic}). The same data augmentation approaches mentioned in the previous paragraph are applied to the synthetic dataset, increasing the complexity of the corresponding learning problem.

While the synthetic dataset is a simplified representation of the real-world respiratory activity, it enables the generation of an arbitrarily large amount of training instances. This synthetic dataset is used to train the 3D-CNN until full convergence without overfitting issues, although on a slightly different---and possibly simpler---learning problem. The small real dataset is ultimately employed solely to \emph{fine-tune} the 3D-CNN previously trained on synthetic data. This is necessary to align it with the true data distribution corresponding to the real-world respiration activity.
Notably, we anticipate that the synthetic data approach has demonstrated remarkable efficacy in mitigating overfitting and consequently improving the generalization results.

It is worth remarking that the selection of the 3D-CNN in this study was guided by preliminary experiments on synthetic videos, which indicated that it is slightly easier to train for this task compared to other architectures of similar complexity, such as Long-term Recurrent Convolutional Networks~\cite{donahue2015long}. Other architectures, including the recently developed Video Vision Transformers~\cite{arnab2021vivit}, may also be viable. However, alternative architectures were not further explored, as the primary challenge lies in data scarcity, which is addressed through data augmentation and synthetic data generation.

\section{Results}
\subsection{Training setup}

A large portion of the dataset was used for training, comprising frames from 25 out of the 30 available subjects, while the remaining 5 were held out as a validation set for performance assessment. The training loss function was set to the Mean Squared Error (MSE) between the reference respiratory frequency $\fR$ extracted as described in Section~\ref{sec:resp_freq_extraction} and the neural network output, while for performance evaluation both the MSE and the coefficient of determination $R^2$ were considered.

Initially, architecture and training hyperparameters were optimized on the synthetic dataset to establish a baseline for comparison, leveraging Bayesian optimization~\cite{bayesianoptml} to tune crucial parameters such as the learning rate, batch size, number of layers and internal channels of the architecture across multiple training instances.
In this investigation, the architecture depicted in Fig.~\ref{Architecture}, along with a learning rate of $10^{-4}$ and a batch size of 32, were identified as favorable among the various configurations explored.

Training on real data was then conducted over 500 epochs, with the best-performing model selected based on validation performance throughout these epochs. In each epoch, the MSE loss was minimized over the entire training dataset, with minibatches extracted in random order from the dataset to promote diversity in learning. After completion of each training epoch, the MSE and the $R^2$ score are computed on the validation dataset to evaluate the model's performance.

The base setup described above was adopted in the following experiments, which were repeated for both the original dataset scaled to resolution $60 \times 80$ (Fig.~\ref{IRframes}) and for the pre-processed dataset (Fig.~\ref{Preprocessed}).

\subsection{Base training}
\begin{figure}[t]
	\centering
	\includegraphics[width=0.8\linewidth]{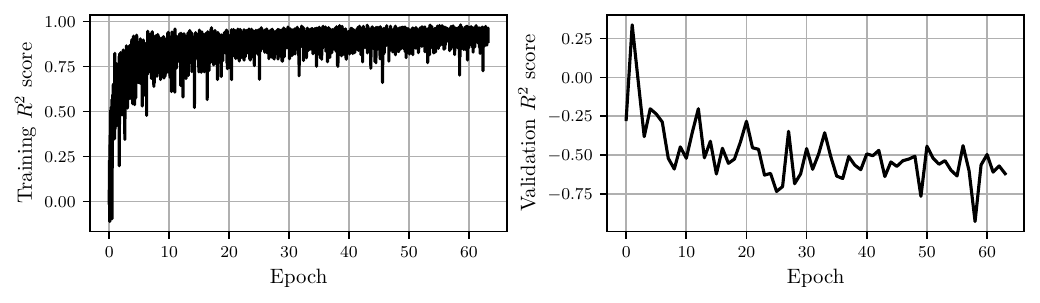}
	\caption{$R^2$ score in training (left) and validation (right) over the training epochs using the base training setup on the unprocessed dataset.}
	\label{Overfit}
\end{figure}

The base training setup described above was initially implemented. However, this approach unveiled notable overfitting challenges, as evidenced by the $R^2$ score converging to 1 on the training dataset, while deteriorating on the validation dataset (see Fig.~\ref{Overfit} and first column block in Table~\ref{results}).
This highlights the need for augmentation strategies to mitigate overfitting and enhance model generalization.

\subsection{Data Augmentation}
Next, data augmentation was introduced to alleviate the overfitting issue and thus potentially enhance results. Data augmentation was implemented using the \texttt{RandAugment} method~\cite{randomaugment}, as implemented in the \texttt{torchvision} library~\cite{torchvision2016} and described in Section~\ref{sec:3DCNN}.
However, \texttt{RandAugment} alone did not lead to performance improvements, as reported in the second column block of Table~\ref{results}.
This outcome suggests that, while standard data augmentation techniques may introduce beneficial variability, they can still fail in an extremely low-data regime, where tailor-made solutions have to be sought for.

\subsection{Transfer Learning from Synthetic Data}
The transfer learning approach from the synthetic dataset described in Section~\ref{sec:3DCNN} was finally adopted to improve model generalization performance.
This approach proved highly effective, largely mitigating overfitting issues and yielding significant performance improvements. This is highlighted in the third column block (Synthetic TL) of Table~\ref{results}, with a validation $R^2$ of 0.488 and 0.521 on the original and pre-processed datasets, respectively.

\begin{figure}[t]
	\centering
	\includegraphics[width=0.8\linewidth]{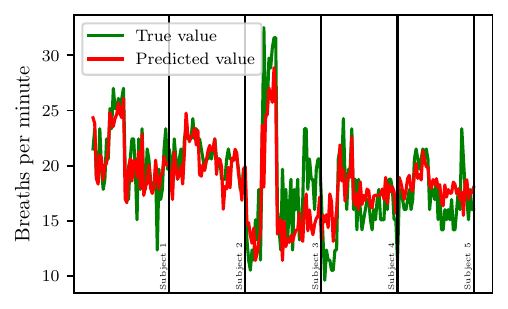}
	\caption{True and predicted values of the model trained on the unprocessed dataset, for all non-overlapping 150-frame segments of the validation split.}
	\label{PredTrue}
\end{figure}

\begin{figure}[t]
	\centering
	\includegraphics[width=0.6\linewidth]{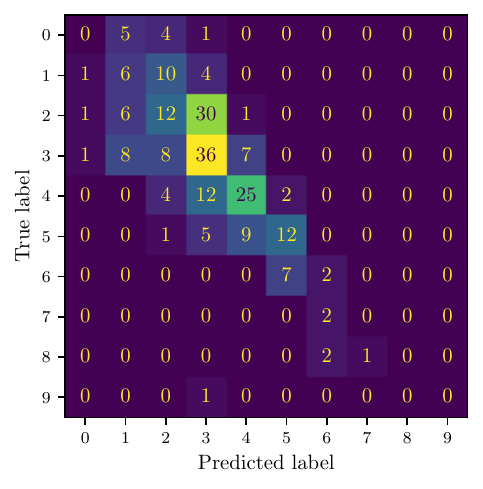}
	\caption{Confusion matrix of the model's predictions on the validation dataset. Labels are obtained by dividing the frequency range into ten equally spaced intervals. The concentration of values along the main diagonal indicates consistent predictive performance.}
	\label{ConfMat}
\end{figure}

Finally, the combination of Transfer Learning from Synthetic Data with Data Augmentation using \texttt{RandAugment} was tested. The achieved results reported in last column block (Augmentation + Synthetic TL) of Table~\ref{results} demonstrate the advantage of this combined approach. Indeed, the validation $R^2$ increases to 0.604/0.614 on the original/pre-processed IR frames.
Interestingly, while Data Augmentation does not lead to any performance improvement when used alone, it provides a substantial performance boost when it is combined with the
Synthetic Data Transfer Learning approach.
In Fig.~\ref{PredTrue}, the predictions of the resulting model on the validation dataset are compared to the actual values. Additionally, Fig.~\ref{ConfMat} displays the confusion matrix for the same validation dataset, with labels derived by dividing the frequency output range into ten equally spaced intervals. Note that the model accurately predicts trends in respiratory frequency, as well as the overall order of magnitude of the signal.

\begin{table}[t]
\centering
\caption{Validation MSE and $R^2$ obtained on the unprocessed and pre-processed IR datasets with the different training strategies. Best results in bold.}
\label{results}
\begin{tabular}{l cc cc cc cc}
\toprule
 & \multicolumn{2}{c}{Base setup} & \multicolumn{2}{c}{Augmentation} & \multicolumn{2}{c}{Synthetic TL} & \multicolumn{2}{c}{Augm.\ + Synthetic TL} \\
\cmidrule(lr){2-3}\cmidrule(lr){4-5}\cmidrule(lr){6-7}\cmidrule(lr){8-9}
Dataset & MSE & $R^2$ & MSE & $R^2$ & MSE & $R^2$ & MSE & $R^2$ \\
\midrule
Unprocessed IR   & 0.752 & 0.045 & 0.789 & 0.000 & 0.403 & 0.488 & \textbf{0.312} & \textbf{0.604} \\
Pre-processed IR & 0.619 & 0.212 & 0.736 & 0.065 & 0.378 & 0.521 & \textbf{0.304} & \textbf{0.614} \\
\bottomrule
\end{tabular}
\end{table}

\begin{figure}[t]
	\centering
	\includegraphics[width=\textwidth]{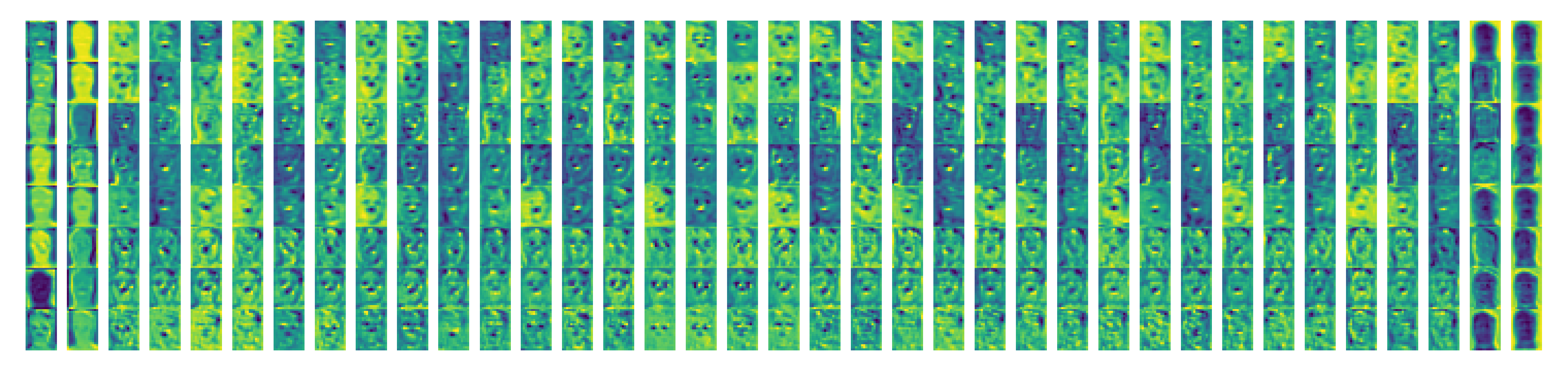}\\[0.5em]
	\includegraphics[width=\textwidth]{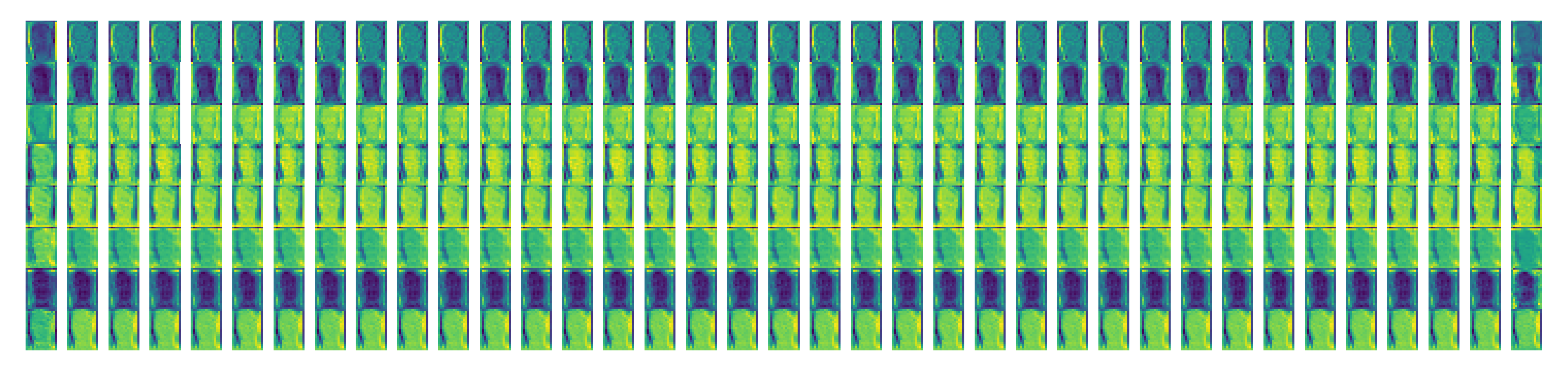}
	\caption{Activations of the second convolutional layer for the trained 3D-CNN (top panel) and for the 3D-CNN at initialization (bottom panel).
	The 8 rows correspond to the channel depth of the layer, and the 37 columns correspond to the reduced time dimension.
	In the trained network, several neurons focus on the nostril region, whereas the untrained network shows little to no response in this area.}
	\label{fig:activations}
\end{figure}

Figure~\ref{fig:activations} shows all the activations of the second convolutional layer for the trained 3D-CNN (top panel) and of the same network at initialization, thus with random weights (bottom panel). Both networks are fed by the same input video, which corresponds to one of the 5 validation subjects. It is evident that, in the trained network, most of the neurons are focused on the nostril region of the face, in line with our expectations.

\begin{figure}[t]
	\centering
	\includegraphics[width=0.8\linewidth]{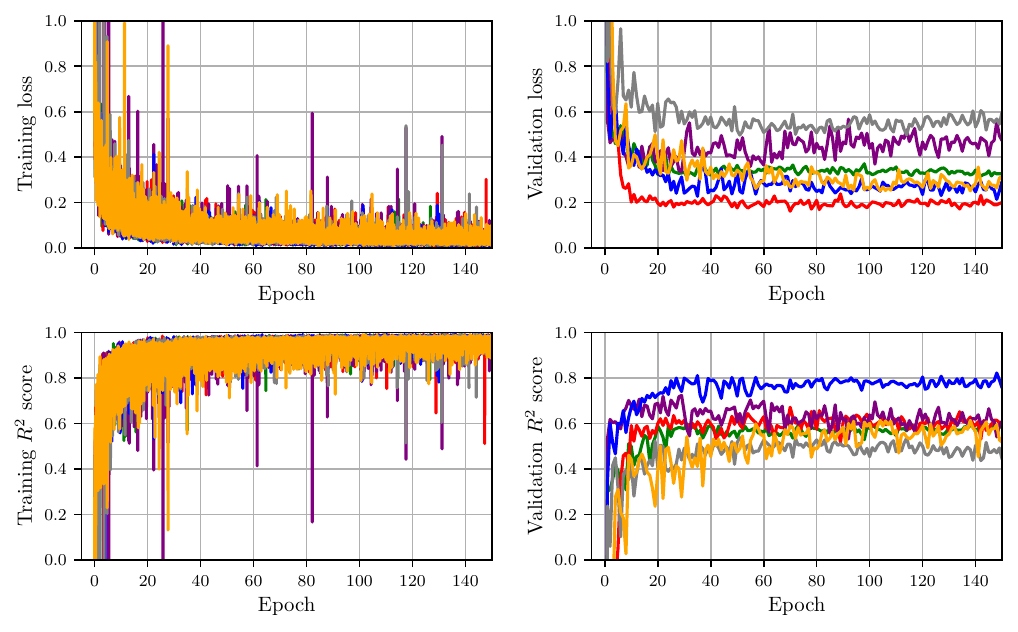}
	\caption{Results of 6-fold cross-validation on the unprocessed dataset, with training performed via augmented synthetic transfer learning.
	The mean validation loss ($0.297$) and mean validation $R^2$ score ($0.663$) across the splits are consistent with the results from the simpler validation setting reported in Table~\ref{results} ($0.312$ and $0.604$, respectively).}
	\label{CrossVal}
\end{figure}

To further validate consistency of the results obtained through the synthetic transfer learning approach, 6-fold cross-validation was conducted on the unprocessed data over all the available subjects, resulting in five additional training/validation splits.
In Fig.~\ref{CrossVal}, the MSE loss and $R^2$ score over the training epochs for the 6 cross-validation splits are shown.
The mean of the validation MSE loss and $R^2$ score over the splits are 0.297 and 0.663, respectively.
This is consistent with the result in the last column of Table~\ref{results}, obtained in a simple validation setting.

\section{Discussion}

In this study, a novel approach for estimating respiratory frequency from thermal signals using deep learning techniques, specifically a 3D-CNN, is introduced.

Compared to traditional approaches, the proposed solution offers an end-to-end framework that eliminates the need for sophisticated pre-processing steps, such as ROI tracking~\cite{fei2009thermistor} or eyes/background removal~\cite{gioia2024thermica}. This holds promise for simplifying existing pipelines for respiratory monitoring.

Traditional methods for respiratory signal extraction from thermal videos typically rely on identifying and tracking specific anatomical landmarks, such as the nostrils or nose. These methods often involve extensive pre-processing to ensure accurate ROI detection and motion compensation. For instance, algorithms utilizing continuous wavelet transforms or Canny edge detectors have shown promise, but are heavily dependent on the quality of ROI detection and tracking \cite{fei2006analysis, abbas2011neonatal, kiashari2020evaluation}.

The need for extensive pre-processing, often involving non-automatic techniques, has been a significant barrier to the practical application of thermography outside controlled laboratory settings. For example, methods requiring manual ROI selection or sophisticated tracking algorithms can be cumbersome and prone to user error \cite{cardone2023facial,sonkusare2021data,gioia2023investigating}. This necessity for strong processing and manual intervention limits the scalability and widespread adoption of thermal imaging technologies in real applications. The presented algorithm, although trained on carefully and ad-hoc pre-processed data, is designed to operate effectively on raw thermal videos. To further validate these claims, both the raw thermal dataset and the pre-processed dataset were used and compared, showing very similar results.
This capability to handle raw data represents a remarkable advancement beyond the state of the art. It opens up numerous possibilities for using thermal imaging in more practical settings, such as remote monitoring, field diagnostics, and real-time surveillance systems, for example, in infant monitoring. The reduced reliance on pre-processing not only streamlines the workflow but also enhances the feasibility of using thermal cameras in diverse and less controlled environments.

Furthermore, the study addresses the common challenge of overfitting in deep learning models, especially when dealing with small datasets. By employing data augmentation strategies and leveraging transfer learning from a synthetic dataset, the generalization capabilities of the model are significantly improved. The combination of these techniques showed a validation $R^2$ score of approximately 0.61, indicating a reliable performance across different validation sets.
It is worthwhile noting that one of the critical challenges in utilizing thermal cameras for data acquisition is the complexity and cost associated with collecting large high-quality datasets, which are usually needed for deep learning approaches. Not only high quality thermal imaging equipment is often expensive, but also the data collection process can be challenging and time-consuming. In this context, the data augmentation strategy employed in this study presents a significant advantage. By generating synthetic data that mimics real thermal video patterns, the training dataset, which included only thirty subjects, was effectively expanded, allowing the model to learn more generalized features and reduce overfitting.

This approach is not only beneficial for our specific application, but also holds promise for a wide range of future studies in thermal imaging. For example, researchers working on facial recognition, stress detection, and other physiological signal monitoring using thermal cameras can adopt similar data augmentation techniques to enhance their models' robustness and performance. The ability to generate high-quality synthetic data reduces the dependency on extensive real-world data collection, thus boosting research ability to propose specific solutions to several research problems and developments in different fields based on thermal imaging.

The potential implications of this study are broad and impactful. The ability to accurately extract respiratory signals from thermal videos has significant applications in healthcare, particularly for continuous and non-intrusive monitoring of patients. For example, this can be especially beneficial in neonatal intensive care units, where traditional sensor attachment can be distressing and potentially harmful to the delicate skin of preterm infants. Moreover, in fields such as psychophysiology and driver monitoring, the methodologies proposed in this paper could provide a non-invasive means of assessing respiratory rate, which is a critical indicator of stress and fatigue.

\section{Limitations and Future Works}
Despite the promising results, several challenges and limitations remain open for future investigations.

While the use of synthetic and augmented data for transfer learning significantly reduces overfitting, these approaches may not fully capture the complexity and variability of real-world respiratory patterns in applications like driver monitoring, where significant head motion is expected.
Additionally, the experimental protocol did not include physical activity, which limited the explored respiratory frequencies to a range of about $9$ to $30$ cycles per minute.

Multiple datasets could be used to assess and possibly improve the robustness of the 3D-CNN approach in those challenging scenarios.
For instance, the multi-modal public domain dataset described in~\cite{taamneh2017multimodal}, which provides thermal/visible-light recordings and physiological signals of human subjects performing distracted driving, could be instrumental to validate the 3D-CNN approach in automotive applications.

Tailor-made data augmentation techniques, such as time-varying rotation/translations of the video stream~\cite{kim2022exploring}, may also be applied to robustify the algorithm against head movements. Furthermore, different respiratory frequencies may be explored by altering the reproduction speed of existing recordings.

Future work will also focus on improving the realism and diversity of synthetic data exploiting, for instance,
3D simulation engines and generative artificial intelligence tools.

While the respiratory frequency is one of the most commonly
used metrics in affective, clinical, and remote monitoring fields,
many other physiological variables are useful for inferring affective states and to perform wellness analysis.
Future investigations will focus on the reconstruction of different aspects of the respiratory signal such as breathing depth and time quotient~\cite{sun2024new}, as well as on the estimation of other physiological signals such as heart rate or blood oxygen levels, to broaden the utility of the presented methodology in healthcare and beyond.
A potential avenue to enhance the robustness of physiological signal reconstruction is the integration of multi-sensor data, combining thermal imaging with other modalities such as visible light or depth cameras.

Finally, the original dataset used in this paper relies on high-quality thermal imaging equipment, which may not be accessible in all settings due to cost and dimension constraints.
Investigating the application of the presented methodologies in real-time monitoring systems based on low-cost thermal cameras could open new possibilities for continuous health monitoring in both clinical and non-clinical settings.

\section{Conclusion}

In conclusion, this study presents a novel deep learning-based approach for extracting respiratory signals from thermal videos.
The proposed 3D-CNN network was able to reconstruct respiratory frequency with an $R^2$ index of approximately $0.61$ for both raw and pre-processed video signals.
By largely eliminating the need for extensive pre-processing and ROI tracking, the presented methodology simplifies the workflow and enhances robustness. The combination of data augmentation and synthetic transfer learning has proven effective in addressing the challenges of small datasets, resulting in reliable performance across different validation sets. The implications of this work are far-reaching, with potential applications in healthcare, psychophysiology, and beyond. Future research will aim to further refine and extend the capabilities of this approach, ensuring its applicability in diverse real-world scenarios.

\section*{Acknowledgments}
The activities of Filippo Pura, Marco Forgione, Arcangelo Merla and Dario Piga have been supported by the Eureka Eurostars project HIRA - Hybrid InfraRed System for Affective Computing, Project id 2335. The activities of Alberto Greco and Federica Gioia have been supported by the Italian Ministry of Education and Research (MUR) through the FoReLab project (Departments of Excellence) and the PNRR---M4C2---Investment 1.3, Partenariato Esteso PE00000013---``FAIR - Future Artificial Intelligence Research''---Spoke 1 ``Human-centered AI''.

\bibliographystyle{unsrtnat}
\bibliography{references}

@article{massaroni2020contactless,
	title={Contactless methods for measuring respiratory rate: A review},
	author={Massaroni, Carlo and Nicolo, Andrea and Sacchetti, Massimo and Schena, Emiliano},
	journal={IEEE Sensors Journal},
	volume={21},
	number={11},
	pages={12821--12839},
	year={2020},
	publisher={IEEE}
}

@article{cardone2023facial,
  title={Facial functional networks during resting state revealed by thermal infrared imaging},
  author={Cardone, Daniela and Cerritelli, Francesco and Chiacchiaretta, Piero and Perpetuini, David and Merla, Arcangelo},
  journal={Physical and Engineering Sciences in Medicine},
  volume={46},
  number={4},
  pages={1573--1588},
  year={2023},
  publisher={Springer}
}

@article{massaroni2019non,
	title={Non-contact monitoring of breathing pattern and respiratory rate via RGB signal measurement},
	author={Massaroni, Carlo and Lo Presti, Daniela and Formica, Domenico and Silvestri, Sergio and Schena, Emiliano},
	journal={Sensors},
	volume={19},
	number={12},
	pages={2758},
	year={2019},
	publisher={MDPI}
}

@article{mateu2019non,
	title={Non-contact infrared-depth camera-based method for respiratory rhythm measurement while driving},
	author={Mateu-Mateus, Marc and Guede-Fernandez, Federico and Garc{\'\i}a-Gonz{\'a}lez, Miguel A and Ramos-Castro, Juan and Fernandez-Chimeno, Mireya},
	journal={IEEE Access},
	volume={7},
	pages={152522--152532},
	year={2019},
	publisher={IEEE}
}

@article{grassmann2016role,
	title={The role of respiratory measures to assess mental load in pilot selection},
	author={Grassmann, Mariel and Vlemincx, Elke and von Leupoldt, Andreas and Van den Bergh, Omer},
	journal={Ergonomics},
	volume={59},
	number={6},
	pages={745--753},
	year={2016},
	publisher={Taylor \& Francis}
}

@article{jakkaew2020non,
	title={Non-contact respiration monitoring and body movements detection for sleep using thermal imaging},
	author={Jakkaew, Prasara and Onoye, Takao},
	journal={Sensors},
	volume={20},
	number={21},
	pages={6307},
	year={2020},
	publisher={MDPI}
}

@article{jagadev2020non,
	title={Non-contact monitoring of human respiration using infrared thermography and machine learning},
	author={Jagadev, Preeti and Giri, Lalat Indu},
	journal={Infrared Physics \& Technology},
	volume={104},
	pages={103117},
	year={2020},
	publisher={Elsevier}
}

@article{kiashari2020evaluation,
	title={Evaluation of driver drowsiness using respiration analysis by thermal imaging on a driving simulator},
	author={Kiashari, Serajeddin Ebrahimian Hadi and Nahvi, Ali and Bakhoda, Hamidreza and Homayounfard, Amirhossein and Tashakori, Masoumeh},
	journal={Multimedia Tools and Applications},
	volume={79},
	pages={17793--17815},
	year={2020},
	publisher={Springer}
}

@article{pereira2018noncontact,
	title={Noncontact monitoring of respiratory rate in newborn infants using thermal imaging},
	author={Pereira, Carina Barbosa and Yu, Xinchi and Goos, Tom and Reiss, Irwin and Orlikowsky, Thorsten and Heimann, Konrad and Venema, Boudewijn and Blazek, Vladimir and Leonhardt, Steffen and Teichmann, Daniel},
	journal={IEEE transactions on Biomedical Engineering},
	volume={66},
	number={4},
	pages={1105--1114},
	year={2018},
	publisher={IEEE}
}

@article{scebba2020multispectral,
	title={Multispectral video fusion for non-contact monitoring of respiratory rate and apnea},
	author={Scebba, Gaetano and Da Poian, Giulia and Karlen, Walter},
	journal={IEEE Transactions on Biomedical Engineering},
	volume={68},
	number={1},
	pages={350--359},
	year={2020},
	publisher={IEEE}
}

@article{stroop1935studies,
  title={Studies of interference in serial verbal reactions.},
  author={Stroop, J Ridley},
  journal={Journal of experimental psychology},
  volume={18},
  number={6},
  pages={643},
  year={1935},
  publisher={Psychological Review Company}
}

@article{sonkusare2021data,
	title={Data-driven analysis of facial thermal responses and multimodal physiological consistency among subjects},
	author={Sonkusare, Saurabh and Breakspear, Michael and Pang, Tianji and Nguyen, Vinh Thai and Frydman, Sascha and Guo, Christine Cong and Aburn, Matthew J},
	journal={Scientific reports},
	volume={11},
	number={1},
	pages={1--12},
	year={2021},
	publisher={Nature Publishing Group}
}

@article{cardone2021automated,
	title={Automated warping procedure for facial thermal imaging based on features identification in the visible domain},
	author={Cardone, Daniela and Spadolini, Edoardo and Perpetuini, David and Filippini, Chiara and Chiarelli, Antonio Maria and Merla, Arcangelo},
	journal={Infrared Physics \& Technology},
	volume={112},
	pages={103595},
	year={2021},
	publisher={Elsevier}
}

@inproceedings{nirkin2018face,
	title={On face segmentation, face swapping, and face perception},
	author={Nirkin, Yuval and Masi, Iacopo and Tuan, Anh Tran and Hassner, Tal and Medioni, Gerard},
	booktitle={2018 13th IEEE International Conference on Automatic Face \& Gesture Recognition (FG 2018)},
	pages={98--105},
	year={2018},
	organization={IEEE}
}

@book{lang2005international,
	title={International affective picture system (IAPS): Affective ratings of pictures and instruction manual},
	author={Lang, Peter J and Bradley, Margaret M and Cuthbert, Bruce N and others},
	year={2005},
	publisher={NIMH, Center for the Study of Emotion \& Attention Gainesville, FL}
}

@inproceedings{gioia2023investigating,
  title={Investigating the Physiology Behind Nose Thermal Response to Stress: A Cross-Mapping Approach},
  author={Gioia, Federica and Nardelli, Mimma and Scilingo, Enzo Pasquale and Greco, Alberto},
  booktitle={Mediterranean Conference on Medical and Biological Engineering and Computing},
  pages={77--85},
  year={2023},
  organization={Springer}
}

@article{ebrahimian2022multi,
	title={Multi-Level Classification of Driver Drowsiness by Simultaneous Analysis of ECG and Respiration Signals Using Deep Neural Networks},
	author={Ebrahimian, Serajeddin and Nahvi, Ali and Tashakori, Masoumeh and Salmanzadeh, Hamed and Mohseni, Omid and Lepp{\"a}nen, Timo},
	journal={International journal of environmental research and public health},
	volume={19},
	number={17},
	pages={10736},
	year={2022},
	publisher={MDPI}
}

@article{nicolo2020importance,
	title={The importance of respiratory rate monitoring: From healthcare to sport and exercise},
	author={Nicol{\`o}, Andrea and Massaroni, Carlo and Schena, Emiliano and Sacchetti, Massimo},
	journal={Sensors},
	volume={20},
	number={21},
	pages={6396},
	year={2020},
	publisher={MDPI}
}

@article{abbas2011neonatal,
	title={Neonatal non-contact respiratory monitoring based on real-time infrared thermography},
	author={Abbas, Abbas K and Heimann, Konrad and Jergus, Katrin and Orlikowsky, Thorsten and Leonhardt, Steffen},
	journal={Biomedical engineering online},
	volume={10},
	number={1},
	pages={1--17},
	year={2011},
	publisher={BioMed Central}
}

@article{scebba2018ir,
	title={IR thermography-based monitoring of respiration phase without image segmentation},
	author={barbosa, K and Rabell, J Esquivelzeta and Del Olmo, P Martin and Haesler, Sebastian},
	journal={Journal of neuroscience methods},
	volume={301},
	pages={1--8},
	year={2018},
	publisher={Elsevier}
}

@article{goodman2019challenges,
	title={Challenges in the diagnosis of paediatric pneumonia in intervention field trials: recommendations from a pneumonia field trial working group},
	author={Goodman, Dina and Crocker, Mary E and Pervaiz, Farhan and McCollum, Eric D and Steenland, Kyle and Simkovich, Suzanne M and Miele, Catherine H and Hammitt, Laura L and Herrera, Phabiola and Zar, Heather J and others},
	journal={The Lancet Respiratory Medicine},
	volume={7},
	number={12},
	pages={1068--1083},
	year={2019},
	publisher={Elsevier}
}

@article{rolfe2019importance,
	title={The importance of respiratory rate monitoring},
	author={Rolfe, Sandy},
	journal={British Journal of Nursing},
	volume={28},
	number={8},
	pages={504--508},
	year={2019},
	publisher={MA Healthcare London}
}

@inproceedings{pereira2017estimation,
	title={Estimation of respiratory rate from thermal videos of preterm infants},
	author={Pereira, Carina Barbosa and Heimann, Konrad and Venema, Boudewijn and Blazek, Vladimir and Czaplik, Michael and Leonhardt, Steffen},
	booktitle={2017 39th Annual International Conference of the IEEE Engineering in Medicine and Biology Society (EMBC)},
	pages={3818--3821},
	year={2017},
	organization={IEEE}
}

@article{barbosa2017estimation,
	title={Estimation of breathing rate in thermal imaging videos: a pilot study on healthy human subjects},
	author={Barbosa Pereira, Carina and Yu, Xinchi and Czaplik, Michael and Blazek, Vladimir and Venema, Boudewijn and Leonhardt, Steffen},
	journal={Journal of clinical monitoring and computing},
	volume={31},
	pages={1241--1254},
	year={2017},
	publisher={Springer}
}

@article{maurya2023non,
	title={Non-contact respiratory rate monitoring using thermal and visible imaging: a pilot study on neonates},
	author={Maurya, Lalit and Zwiggelaar, Reyer and Chawla, Deepak and Mahapatra, Prasant},
	journal={Journal of Clinical Monitoring and Computing},
	volume={37},
	number={3},
	pages={815--828},
	year={2023},
	publisher={Springer}
}

@article{pramanik2015neonatal,
	title={Neonatal respiratory distress: a practical approach to its diagnosis and management},
	author={Pramanik, Arun K and Rangaswamy, Nandeesh and Gates, Thomas},
	journal={Pediatric Clinics},
	volume={62},
	number={2},
	pages={453--469},
	year={2015},
	publisher={Elsevier}
}

@inproceedings{cho2017deepbreath,
	title={DeepBreath: Deep learning of breathing patterns for automatic stress recognition using low-cost thermal imaging in unconstrained settings},
	author={Cho, Youngjun and Bianchi-Berthouze, Nadia and Julier, Simon J},
	booktitle={2017 Seventh international conference on affective computing and intelligent interaction (acii)},
	pages={456--463},
	year={2017},
	organization={IEEE}
}

@article{flenady2017accurate,
	title={Accurate respiratory rates count: So should you!},
	author={Flenady, Tracy and Dwyer, Trudy and Applegarth, Judith},
	journal={Australasian emergency nursing journal},
	volume={20},
	number={1},
	pages={45--47},
	year={2017},
	publisher={Elsevier}
}

@article{elliott2016respiratory,
	title={Why is respiratory rate the neglected vital sign? A narrative review},
	author={Elliott, Malcolm},
	journal={Int Arch Nurs Health Care},
	volume={2},
	number={3},
	pages={050},
	year={2016}
}

@article{nicolo2020respiratory,
	title={Respiratory frequency as a marker of physical effort during high-intensity interval training in soccer players},
	author={Nicol{\`o}, Andrea and Montini, Marco and Girardi, Michele and Felici, Francesco and Bazzucchi, Ilenia and Sacchetti, Massimo},
	journal={International journal of sports physiology and performance},
	volume={15},
	number={1},
	pages={73--80},
	year={2020},
	publisher={Human Kinetics}
}

@article{loughlin2018respiratory,
	title={Respiratory rate: The forgotten vital sign—Make it count!},
	author={Loughlin, Patrick C and Sebat, Frank and Kellett, John G},
	journal={Joint Commission journal on quality and patient safety},
	volume={44},
	number={8},
	pages={494--499},
	year={2018},
	publisher={Elsevier}
}

@article{nicolo2017respiratory,
	title={Respiratory frequency during exercise: the neglected physiological measure},
	author={Nicol{\`o}, Andrea and Massaroni, Carlo and Passfield, Louis},
	journal={Frontiers in physiology},
	volume={8},
	pages={922},
	year={2017},
	publisher={Frontiers Media SA}
}

@article{kumar2020continuous,
	title={Continuous vital sign analysis for predicting and preventing neonatal diseases in the twenty-first century: big data to the forefront},
	author={Kumar, Navin and Akangire, Gangaram and Sullivan, Brynne and Fairchild, Karen and Sampath, Venkatesh},
	journal={Pediatric research},
	volume={87},
	number={2},
	pages={210--220},
	year={2020},
	publisher={Nature Publishing Group US New York}
}

@article{hu2017synergetic,
	title={Synergetic use of thermal and visible imaging techniques for contactless and unobtrusive breathing measurement},
	author={Hu, Meng-Han and Zhai, Guang-Tao and Li, Duo and Fan, Ye-Zhao and Chen, Xiao-Hui and Yang, Xiao-Kang},
	journal={Journal of biomedical optics},
	volume={22},
	number={3},
	pages={036006--036006},
	year={2017},
	publisher={Society of Photo-Optical Instrumentation Engineers}
}

@article{siddiqui2021non,
	title={Non-invasive driver drowsiness detection system},
	author={Siddiqui, Hafeez Ur Rehman and Saleem, Adil Ali and Brown, Robert and Bademci, Bahattin and Lee, Ernesto and Rustam, Furqan and Dudley, Sandra},
	journal={Sensors},
	volume={21},
	number={14},
	pages={4833},
	year={2021},
	publisher={MDPI}
}

@article{3dcnnactionrecognition,
  author={Ji, Shuiwang and Xu, Wei and Yang, Ming and Yu, Kai},
  journal={IEEE Transactions on Pattern Analysis and Machine Intelligence}, 
  title={{3D Convolutional Neural Networks for Human Action Recognition}}, 
  year={2013},
  volume={35},
  number={1},
  pages={221-231},
  doi={10.1109/TPAMI.2012.59}
}

@inproceedings{tran2015learning,
  title={Learning spatiotemporal features with 3d convolutional networks},
  author={Tran, Du and Bourdev, Lubomir and Fergus, Rob and Torresani, Lorenzo and Paluri, Manohar},
  booktitle={Proceedings of the IEEE international conference on computer vision},
  pages={4489--4497},
  year={2015}
}

@misc{bayesianoptml,
      title={Practical Bayesian Optimization of Machine Learning Algorithms}, 
      author={Jasper Snoek and Hugo Larochelle and Ryan P. Adams},
      year={2012},
      eprint={1206.2944},
      archivePrefix={arXiv},
      primaryClass={stat.ML}
}

@article{xu2023comprehensive,
  title={A comprehensive survey of image augmentation techniques for deep learning},
  author={Xu, Mingle and Yoon, Sook and Fuentes, Alvaro and Park, Dong Sun},
  journal={Pattern Recognition},
  volume={137},
  pages={109347},
  year={2023},
  publisher={Elsevier}
}

@article{randomaugment,
  author       = {Ekin D. Cubuk and
                  Barret Zoph and
                  Jonathon Shlens and
                  Quoc V. Le},
  title        = {RandAugment: Practical data augmentation with no separate search},
  journal      = {CoRR},
  volume       = {abs/1909.13719},
  year         = {2019},
  url          = {http://arxiv.org/abs/1909.13719},
  eprinttype    = {arXiv},
  eprint       = {1909.13719},
  bibsource    = {dblp computer science bibliography, https://dblp.org}
}

@unpublished{torchvision2016, 
    title        = {TorchVision: PyTorch's Computer Vision library},
    author       = {TorchVision maintainers and contributors},
    year         = 2016,
    journal      = {GitHub repository},
    publisher    = {GitHub},
    howpublished = {\url{https://github.com/pytorch/vision}}
}

@article{kim2022transferable,
  title={How transferable are video representations based on synthetic data?},
  author={Kim, Yo-whan and Mishra, Samarth and Jin, SouYoung and Panda, Rameswar and Kuehne, Hilde and Karlinsky, Leonid and Saligrama, Venkatesh and Saenko, Kate and Oliva, Aude and Feris, Rogerio},
  journal={Advances in Neural Information Processing Systems},
  volume={35},
  pages={35710--35723},
  year={2022}
}

@article{jimenez2018k,
  title={K deep: protein--ligand absolute binding affinity prediction via 3d-convolutional neural networks},
  author={Jim{\'e}nez, Jos{\'e} and Skalic, Miha and Martinez-Rosell, Gerard and De Fabritiis, Gianni},
  journal={Journal of chemical information and modeling},
  volume={58},
  number={2},
  pages={287--296},
  year={2018},
  publisher={ACS Publications}
}

@inproceedings{fei2006analysis,
  title={Analysis of breathing air flow patterns in thermal imaging},
  author={Fei, Jin and Pavlidis, Ioannis},
  booktitle={2006 International Conference of the IEEE Engineering in Medicine and Biology Society},
  pages={946--952},
  year={2006},
  organization={IEEE}
}

@article{Jones2012,
  title={Standard Procedures For Infrared Imaging in Medicine},
  author={Jones, B.F. and Plassmann, P. and Ring, E.F.J. and Ammer, K. and Shterenshis, M. and Chong, A. and Lindberg, L. and Diakides, N.A.},
  journal={IEEE Transactions on Medical Imaging},
  year={2012}
}

@article{gioia2024thermica,
  title={ThermICA: Novel Approach for a Multivariate Analysis of Facial Thermal Responses},
  author={Gioia, Federica and Greco, Alberto and Callara, Alejandro Luis and Vanello, Nicola and Scilingo, Enzo Pasquale and Citi, Luca},
  journal={IEEE Transactions on Biomedical Engineering},
  year={2024},
  publisher={IEEE}
}

@article{taamneh2017multimodal,
  title={A multimodal dataset for various forms of distracted driving},
  author={Taamneh, Salah and Tsiamyrtzis, Panagiotis and Dcosta, Malcolm and Buddharaju, Pradeep and Khatri, Ashik and Manser, Michael and Ferris, Thomas and Wunderlich, Robert and Pavlidis, Ioannis},
  journal={Scientific data},
  volume={4},
  number={1},
  pages={1--21},
  year={2017},
  publisher={Nature Publishing Group}
}

@article{sun2024new,
  title={A New Look At Breathing for Affective Studies},
  author={Sun, Nanfei and Pavlidis, Ioannis},
  journal={IEEE Transactions on Affective Computing},
  number={01},
  pages={1--18},
  year={2024},
  publisher={IEEE Computer Society}
}

@article{fei2009thermistor,
  title={Thermistor at a distance: unobtrusive measurement of breathing},
  author={Fei, Jin and Pavlidis, Ioannis},
  journal={IEEE transactions on biomedical engineering},
  volume={57},
  number={4},
  pages={988--998},
  year={2009},
  publisher={IEEE}
}

@article{kim2022exploring,
  title={Exploring temporally dynamic data augmentation for video recognition},
  author={Kim, Taeoh and Kim, Jinhyung and Shim, Minho and Yun, Sangdoo and Kang, Myunggu and Wee, Dongyoon and Lee, Sangyoun},
  journal={arXiv preprint arXiv:2206.15015},
  year={2022}
}

@inproceedings{donahue2015long,
  title={Long-term recurrent convolutional networks for visual recognition and description},
  author={Donahue, Jeffrey and Anne Hendricks, Lisa and Guadarrama, Sergio and Rohrbach, Marcus and Venugopalan, Subhashini and Saenko, Kate and Darrell, Trevor},
  booktitle={Proceedings of the IEEE conference on computer vision and pattern recognition},
  pages={2625--2634},
  year={2015}
}

@inproceedings{arnab2021vivit,
  title={Vivit: A video vision transformer},
  author={Arnab, Anurag and Dehghani, Mostafa and Heigold, Georg and Sun, Chen and Lu{\v{c}}i{\'c}, Mario and Schmid, Cordelia},
  booktitle={Proceedings of the IEEE/CVF international conference on computer vision},
  pages={6836--6846},
  year={2021}
}

\end{document}